\pdfoutput=1

\documentclass[11pt]{article}

\usepackage[final]{coling}

\usepackage{times}
\usepackage{latexsym}

\usepackage[T1]{fontenc}

\usepackage[utf8]{inputenc}

\usepackage{microtype}

\usepackage{inconsolata}

\usepackage{graphicx}

\usepackage{multirow}
\usepackage{dsfont}
\usepackage{amsmath, amsthm, mathtools, amssymb}
\usepackage{svg}
\usepackage{amsfonts}
\usepackage[shortlabels, inline]{enumitem}
\usepackage{tikz}
\usepackage{collcell}
\usepackage{graphicx}
\usepackage{caption}
\usepackage{listings}
\usepackage{wrapfig}
\usepackage[export]{adjustbox}
\usepackage{colortbl}
\usepackage{scalerel}
\usepackage{ulem} 

\definecolor{dkgreen}{rgb}{0,0.6,0}
\definecolor{gray}{rgb}{0.5,0.5,0.5}
\definecolor{mauve}{rgb}{0.58,0,0.82}
\definecolor{lightgray}{gray}{0.9}
\title{\texttt{LAW}: Legal Agentic Workflows for Custody and Fund Services Contracts}

\author{
    \textbf{William Watson\thanks{Equal Contribution}}, 
    \textbf{Nicole Cho\footnote[1]{Equal Contribution}},
    \\
    \textbf{Nishan Srishankar\footnote[1]{Equal Contribution}},
    \textbf{Zhen Zeng}, 
    \textbf{Lucas Cecchi}, 
    \textbf{Daniel Scott}, 
    \\
    \textbf{Suchetha Siddagangappa}, 
    \textbf{Rachneet Kaur},
    \textbf{Tucker Balch}, 
    \textbf{Manuela Veloso} 
    \\
    J.P.~Morgan AI Research 
    \\ 
    New York, New York, USA
    \\
    \texttt{nicole.cho@jpmorgan.com}
}

\begin{document}
\maketitle
\begin{abstract}
Legal contracts in the custody and fund services domain govern critical aspects such as key provider responsibilities, fee schedules, and indemnification rights. However, it is challenging for an off-the-shelf Large Language Model (LLM) to ingest these contracts due to the lengthy unstructured streams of text, limited LLM context windows, and complex legal jargon. To address these challenges, we introduce \texttt{LAW} (Legal Agentic Workflows for Custody and Fund Services Contracts). \texttt{LAW} features a modular design that responds to user queries by orchestrating a suite of domain-specific tools and text agents. Our experiments demonstrate that \texttt{LAW}, by integrating multiple specialized agents and tools, significantly outperforms the baseline. \texttt{LAW} excels particularly in complex tasks such as calculating a contract's termination date, surpassing the baseline by 92.9\% points. Furthermore, \texttt{LAW} offers a cost-effective alternative to traditional fine-tuned legal LLMs by leveraging reusable, domain-specific tools.
\end{abstract}

\begin{figure*}[t!]
  \centering
  \includegraphics[clip, trim=1.15cm 1.61cm 1.2cm 1.0cm, width=\textwidth]{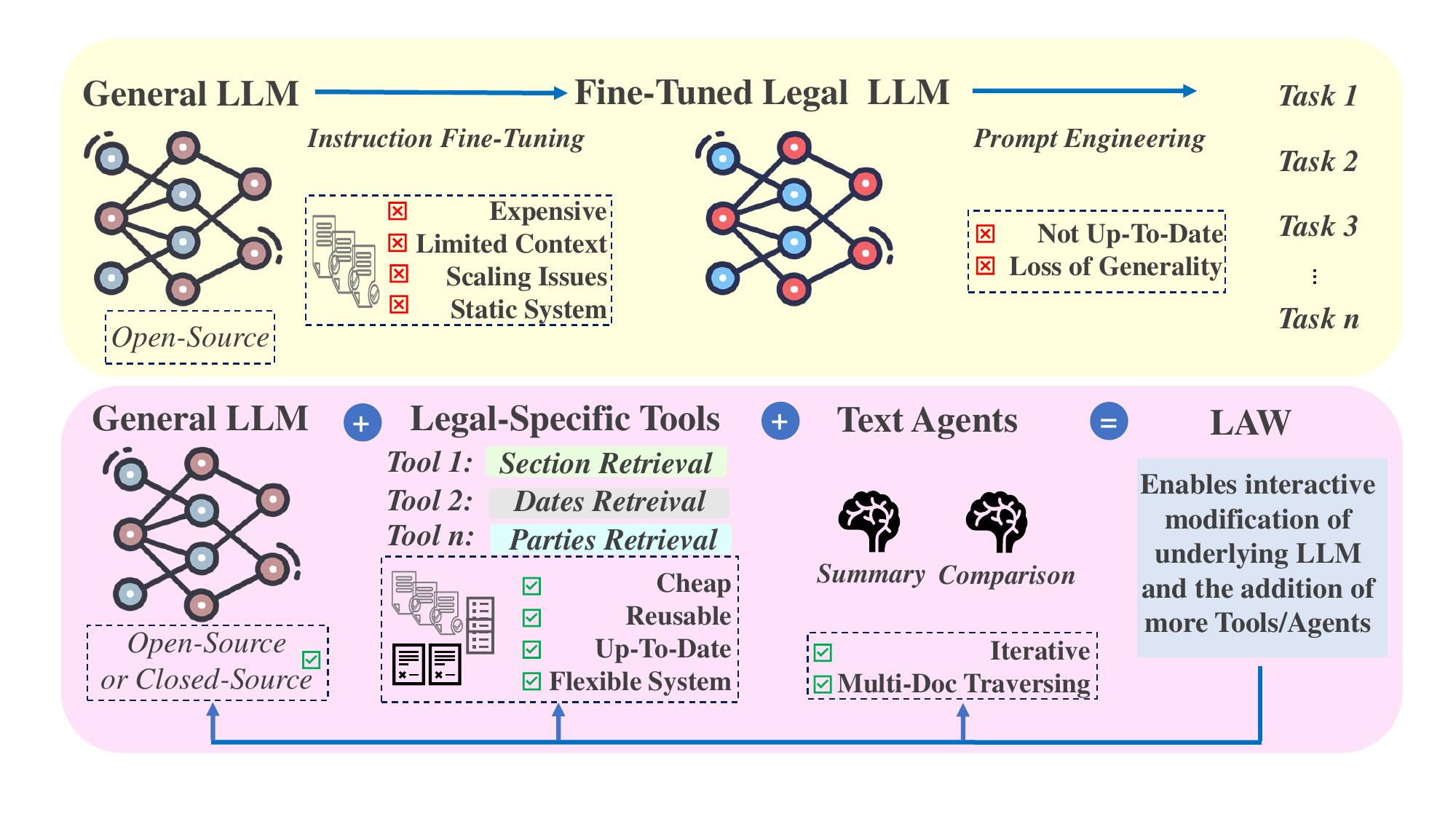}
  \caption{Comparing the traditional method of fine-tuning a legal LLM vs. \texttt{LAW} (Legal Agentic Workflows). Fine-tuning involves labeling contracts for a highly customized pipeline supported by open-source LLMs, which results in limited context, scale, or flexibility where coaxing additional information out of the model would require further tuning. The ensuing prompt engineering on the fine-tuned legal LLM also exacerbates the model's loss of generality. In contrast, \texttt{LAW} can operate on both closed-source or open-source LLMs and is equipped with legal domain-specific tools.  These tools are cheaper to construct, reusable, simpler to construct, and incorporates recent data.  The general LLM's orchestration of these tools along with the text agents engenders \texttt{LAW}, a highly interactive agentic system that also enables the addition of more tools and agents.}
  \label{fig:teaser}
\end{figure*}

\section{Introduction}

While the advancement of Large Language Models (LLMs) demonstrates great potential for a myriad of use-cases in Document AI and Natural Language Processing (NLP) \citep{minaee2024largelanguagemodelssurvey}, the domain of legal contracts poses unique challenges. The necessity for models to comprehend long, multi-document context windows and dense legal jargon engenders the intellectual pursuit to construct a legal domain-specific LLM. Certain studies have empirically investigated this motivation such as comparing the zero-shot performance of general-purpose LLMs on legal texts
\citep{jayakumar-etal-2023-large} or fine-tuning LLMs under the Federated-Learning setting \citep{yue2024fedjudgefederatedlegallarge}. 
Similarly, \citet{colombo2024saullm7bpioneeringlargelanguage} trained \texttt{SaulLM-7B} on an English legal corpus, leveraging the \texttt{Mistral-7B} architecture \citep{jiang2023mistral7b}. While these developments are promising, legal contracts are highly varied not only in terms of semantics but also accessibility. Therefore, compared to the computational cost, the usage of a fine-tuned legal LLM can be very limited in practice (Figure~\ref{fig:teaser}). Thus, we propose \texttt{LAW}, a legal agentic workflow framework, that uses a code generation agent to orchestrate reusable tools, that can be leveraged for a variety of different contracts. Moreover, our framework can be generalized across different types of queries. 
Instead of relying solely on a fine-tuned LLM to solve highly complex tasks, \texttt{LAW} leverages a suite of specialized legal domain-specific tools, and a robust orchestration framework built on top of the FlowMind framework proposed by \citet{10.1145/3604237.3626908}. \texttt{LAW}'s reusable tools are designed to tackle distinct tasks such as contract retrieval. Our tools are rigorously guardrailed through unit tests that map their failure modes, ensuring a comprehensive understanding of their operational limits. By utilizing this method, \texttt{LAW} focuses on selectively applying the appropriate tools and text agents for a task, thereby optimizing the problem-solving process and delivering accurate and reliable responses.

\paragraph{Contributions}
We empirically prove the optimized performance of \texttt{LAW} for complex legal tasks. 
Overall, our contributions are three-fold:
\begin{itemize}[noitemsep, leftmargin=*, topsep=0pt, partopsep=0pt, label={\tiny\raisebox{0.5ex}{$\blacktriangleright$}}]
\item We propose \texttt{LAW}, a novel approach to interacting with financial-legal contracts, utilizing reusable legal domain-specific tools and text agents that addresses practical constraints - specifically legal dataset accessibility, scalability, and cost. Our system that can allow both lay-people, and domain experts to query information from complicated legal documents.
\item \texttt{LAW} significantly outperforms the baseline, achieving up to 92.9\% accuracy gains across a range of queries, from direct retrieval to multi-hop reasoning.
\item \texttt{LAW} is the first legal agentic workflow system encompassing 23 years of regulatory contracts for the entire scope of public funds pursuant to the Investment Company Act of 1940. \texttt{LAW} can perform retrieval and analytical tasks that require an understanding over multiple documents, and each document contains many pages.

\end{itemize}

\section{Related Works}

\paragraph{LLMs in NLP}
LLMs such as Llama 2 \citep{touvron2023llama}, PaLM \citep{chowdhery2023palm}, GPT-3 \citep{brown2020language}, GPT-4 \citep{achiam2023gpt}, and Vicuna-13B \citep{vicuna2023}, have revolutionized NLP in many aspects. Their capabilities provide a foundation for more specialized adaptations, such as InstructGPT \citep{ouyang2022training}, which demonstrates how fine-tuning GPT models with human feedback can significantly enhance their alignment with user intent. Despite their impressive capabilities, LLMs face challenges like hallucination, outdated knowledge, and untraceable reasoning processes. Retrieval-Augmented Generation (RAG) \citep{lewis2020retrieval, siriwardhana2023improving, lin2023ra, gao2023retrieval} has emerged as a promising solution by effectively merging LLMs' intrinsic knowledge with external databases, enhancing both accuracy and reliability of generated content for knowledge-intensive tasks.

\paragraph{Domain-Specific Tools}
FlowMind \citep{10.1145/3604237.3626908} introduces a generic prompt recipe that employs reliable Application Programming Interfaces (APIs) to ground LLM reasoning through the usage of tools. Additionally, HiddenTables \citep{watson-etal-2023-hiddentables} constructed an agentic system designed to enhance interactions with tabular data. 
\citet{chen2023program} and \citet{gao2023pal} explored how models can generate not only coherent text but also executable code snippets based on user queries. ToolFormer \citep{schick2023toolformer} and REACT \citep{yao2023react}, which are designed to enhance the model's interaction with external databases and software, helped LLMs access a wider range of resources, improving their ability to answer queries that required specialized knowledge. \citet{10.1145/3383455.3422520} demonstrated an end-to-end pipeline for financial extraction and transcription of tabular content from images.
Furthermore, adaptations in Text-to-SQL \citep{rajkumar2022evaluating} methods for transforming natural language queries into database-readable commands show promise in streamlining document analysis tasks. CodeAct \citep{wang2024executablecodeactionselicit} demonstrated that executable Python code can unify LLM agents' actions in a single action space, allowing for dynamic adjustments and new policies based on multi-turn interactions. 
Furthermore, LLMs in financial intelligence has evolved from traditional knowledge-graph and database approaches to domain-specific LLMs, though these face challenges with costs and accuracy, motivating the development of more sophisticated architectures \citep{10.1145/3383455.3422520, watson2024thingbadquestionh4r, 10.1145/3677052.3698597}.

\paragraph{Legal LLMs}
 \texttt{SaulLM-7B} \citep{colombo2024saullm7bpioneeringlargelanguage}, based on \texttt{Mistral-7B}, is specifically designed for legal text comprehension and generation. Trained on an extensive English legal corpus, it shows state-of-the-art capabilities in processing legal documents using instruction fine-tuning. \citet{jayakumar-etal-2023-large} explored the zero-shot capabilities of general-purpose LLMs such as ChatGPT-3.5, LLaMA2-70b, and Falcon-180B on contract provision classification, noting their lower F1 scores compared to smaller, legal-specific fine-tuned models. \citet{yue2024fedjudgefederatedlegallarge} presents FedJudge, the inaugural Federated Legal LLM framework, optimizing performance with minimal parameter updates during federated learning. Additionally, \citet{fei2024internlmlawopensourcechinese} introduces InternLM-Law, tailored for diverse legal inquiries related to Chinese laws. 
 \citet{trautmann2022legalpromptengineeringmultilingual} assesses zero-shot Legal Prompt Engineering (LPE) for processing complex legal documents in multiple languages, focusing on legal judgment prediction tasks. Finally, \citet{roegiest-etal-2023-questions} examines the potential of LLMs to generate structured answers to legal questions, specifically in multiple-choice formats.

\paragraph{Evaluation Frameworks in Legal Environments}
\citet{chen2021evaluating} and \citet{nye2021work} provide insight into the performance of LLMs in executing complex tasks.
Their methodologies for assessing the accuracy and transparency of model outputs could be vital for deploying LLMs in legal settings where precision and accountability are crucial. Moreover, 
\citet{liang2023code} offers frameworks for ensuring that LLM operations adhere to legal and ethical standards.
While the existing literature underscores significant advancements in legal LLM applications, \texttt{LAW}'s modular design employs an orchestrator agent integrating reusable tools for legal domain-specific tasks, marking a significant evolution from previous models.

\section{Data Sourcing \& Ingestion}\label{sec:data_source}

\paragraph{EDGAR} 
For our dataset, we procure contracts from EDGAR (Electronic Data Gathering, Analysis, and Retrieval), the U.S. SEC's (Securities and Exchange Commission)\footnote{https://www.sec.gov/}  database of regulatory filings.  23 years of filings are available in the omnibus filing 485BPOS which houses 2.7 million exhibits.  From these, we procure 17,831 legal contracts (Appendix~\ref{app:understanding_contracts}).

\paragraph{Form 485BPOS}
Form 485BPOS is a post-effective amendment filed by all investment companies governed by the US Investment Company Act of 1940. 
These investment companies, colloquially dubbed \textit{'40Act funds}, are mandated to file Form N-1A or Form 485BPOS, pursuant to Securities Act Rule 485(b) \citep{edgar}. We choose the legal contracts housed in Form 485BPOS omnibus filing as they capture the entire universe of all '40Act funds and account for a non-trivial (14\%) of EDGAR filings.

\paragraph{Ingesting Contracts} 
Contracts are difficult to directly ingest due to inconsistent reporting in EDGAR. Moreover, the SEC only allows a maximum throughput of 10 reports/second - this limitation necessitates the need to bring our data on-premise as EDGAR is not accessible at scale. 
In summary, we ingest a total of 22~GB of data on-premise within our knowledge base through a myriad of techniques such as:
\begin{itemize}[noitemsep, leftmargin=*, topsep=0pt, partopsep=0pt, label={\tiny\raisebox{0.5ex}{$\blacktriangleright$}}]
    \item \textbf{Scalable Procurement:} We ingest at a rate of 112 documents/second - 6.7 hours were spent in terms of sequential processing.  These contracts are not individually searchable on EDGAR; our knowledge base enables individual search.
    \item \textbf{AI Metadata Tagging and Search:} Each section is made searchable via title recognition algorithms, alongside contextual and visual cues to intelligently chunk each contract for precise retrieval within our distributed hybrid search.
\end{itemize}

\section{Tools}\label{sec:tools}
We develop legal domain-specific tools that each undertakes a specialized task. These tools enable re-usability across varying contracts in our dataset; moreover, additional tools can be added at any stage of \texttt{LAW}'s development.

\subsection{Tools for Direct Extraction}

\paragraph{Tool for Extracting Dates}\label{sec:extract_dates}
Contracts house different types of dates such as the contract's \texttt{Effective Date} (when the current contract is effective), \texttt{Master Date} (when the master/original contract was effective), and \texttt{Dated Date} (when the current contract was signed). Our tool distinguishes these three types. Detection and extraction of dates is achieved via RoBERTa span detection \citep{liu2019robertarobustlyoptimizedbert}, HTML parsing with BeautifulSoup4\footnote{\url{https://www.crummy.com/software/BeautifulSoup/}}, and regular expression heuristics. Then, our tool standardizes all extracted dates to the \texttt{DD/MM/YYYY} format.

\paragraph{Tool for Extracting Parties}
This tool's objective is to find and extract the associated parties involved in signing the contract. These include the trust of funds and the custodian bank. Filing 485BPOS in EDGAR contains metadata about a subset of the involved parties for certain contracts. This is because the filing metadata only pertains to the specific legal entity making the EDGAR submission, rather than encompassing the full breadth of an investment manager's fund offerings and related parties. 
We use this as a guide to train our system's understanding of the full scope of contracts.
We implement fuzzy matching to search for these parties in the contracts. The custom fuzzy matching built on top of RapidFuzz\footnote{\url{https://github.com/rapidfuzz/RapidFuzz}} aims to mitigate issues that may arise from stylistic differences in names such as special character usage, capitalization, and differing naming conventions. Additionally, we mitigate issues with some names being substrings of others by searching sequentially in order of increasing name length and removing found parties.

\subsection{Tools for Multi-Hop Reasoning}\label{sec:multihop}
\paragraph{Tool to Calculate Contract Lifecycle}
Contracts typically have a lifecycle during which the provisions are enforced. 
This tool aims to calculate the termination date of the contract's lifecycle. It uses our existing tool for dates to extract the effective date of the contracts. Next, it searches for the contract's duration or the termination date. If the contract mentions the duration (e.g. \texttt{3 years}), the tool translates the text into a numerical value. Finally, this numerical value is added to the effective date to generate a termination date.

\paragraph{Tool to Retrieve Master Contract}
This tool's goal is to differentiate between a \texttt{master} and an \texttt{amendment} agreement. Master agreements refer to the original contract that outlines all aspects of the relationship between the fund and the custodian bank. An amendment refers to contracts that amend the master or any subsequent amendments - amendments are typically less detailed as they can amend a single word. 
Our tool classifies and retrieves the master contract by comparing the extracted \textit{effective date} with the \textit{master effective date} using the tool for dates; if equal, the contract is considered to be the master. 
If determined to be an amendment, the tool searches for the master by matching the dates and parties. 

\paragraph{Tool to Label Section Titles}\label{sec:section_title}
Contracts are semantically structured and hierarchical in the construction of their clauses. Each clause holds detailed knowledge regarding terminology such as indemnification, force majure, and termination. Therefore, when queries about particular terms arise, directly retrieving the relevant clause or section is far more efficient than reviewing the entire contract indiscriminately.
However, parsing contracts into distinct semantic sections for effective retrieval presents significant challenges. Many contracts lack explicit section declarations and the language across different sections can be highly similar. 
To address this challenge, we employed a fine-tuned \texttt{t5-large} \citep{2020t5} model trained to classify paragraphs into one of 20 potential section labels. These labels cover a broad spectrum of typical clauses found in contracts (Appendix~\ref{app:clauses}).
Our training dataset is comprised of 1,500 paragraphs per title, systematically collected from a variety of contracts to ensure diverse linguistic representations. As an alternative solution, we trained a \texttt{t5-base} model for title generation instead of classification. Both section title classification and generation models perform similarly. (Appendix~\ref{app:section_accuracy}) outlines the performance and training parameters.
Section titles are then used for section search and retrieval as described in \S\ref{subsec:infra_opensearch}.

\begin{figure*}[t]
\includegraphics[clip, trim=4.5cm 6.0cm 4.5cm 4.5cm,width=\textwidth]{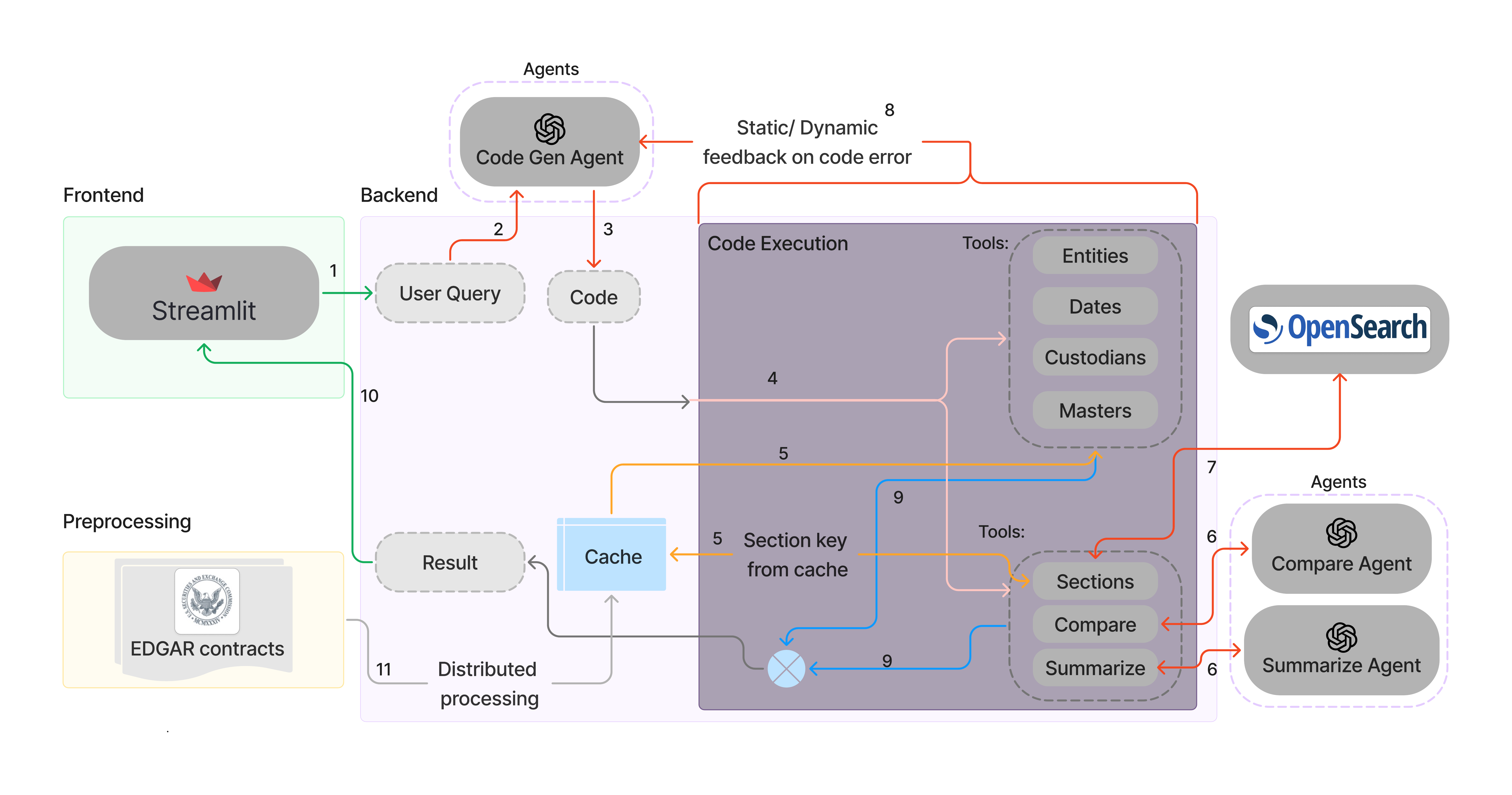}
\captionof{figure}{System Overview of \textbf{\texttt{LAW}}. 
\textbf{(1)} User query input on front-end~(\S\ref{subsec:infra_question_templates});
\textbf{(2)} Query manipulation and custom modification added to prompt and sent to the code generation agent~(\S\ref{subsec:code_genagent});
\textbf{(3)} Chat completion return from the code generation agent;
\textbf{(4)} Execution of backend API tools~(\S\ref{sec:tools});
\textbf{(5)} Tool retrieval of information from internal cache~(\S\ref{subsec:infra_caching});
\textbf{(6)} Calls to text agents~(\S\ref{sec:text_agents});
\textbf{(7)} Calls to multi-node OpenSearch cluster for text retrieval~(\S\ref{subsec:infra_opensearch}); 
\textbf{(8)} Feedback on code runs back to the code generation agent in case of failure;
\textbf{(9)} Concatenation of final output;
\textbf{(10)} Final text output is rendered on the UI;
\textbf{(11)} EDGAR contracts undergo continuous, offline, distributed processing to update our internal cache and OpenSearch systems~(\S\ref{sec:data_source}).
}
\label{fig:system_overview}
\end{figure*}

\section{Text Agents}\label{sec:text_agents}
\paragraph{Summary Agent}
\label{subsec:agents_summarization_agent}
The summary agent aims to provide a useful summary of legal clauses. Our prompts enable the agent to focus  on identifying and preserving key terms such as entity names and dates.
A key challenge in summarizing relates to sections that exceed an LLM's context window. 
Legal contracts can be very long, averaging around 27K $\pm$ 51K tokens when encoded by \texttt{tiktoken}.
Therefore, if the input text and prompt exceed the 16K token limit of \texttt{gpt-3.5-turbo}, the text is split into 8K token chunks. These chunks are processed in parallel by separate sub-agents, with the output concatenated into a final summary.

\paragraph{Comparison Agent}
The comparison agent's purpose is to understand how particular clauses are different across time or entities. With a similar base prompt as the summary agent, it compares two bodies of text.
The agent chronologically sorts the sections from different contracts and, in parallel, compares each pairwise set of sections in the list.
For example, given a list of contracts' sections $\mathcal{L}=[s_0,s_1,...,s_n]$, where $s_i$ is an individual section, it performs $compare(s_i,s_{i+1})\,\forall\, s_i \in \mathcal{L}, i\neq~n$. 
To handle large bodies of text, the agent also performs repeated summaries on each section $s_i$ to condense the body to a manageable size. The summarized sections are then passed to the comparison agent.

\section{Engineering Infrastructure}
\label{sec:eng_infra}
The system overview for \texttt{LAW} is illustrated in Figure~\ref{fig:system_overview}. In addition to the tools described previously, it also uses the following modules:
\paragraph{User Query}
\label{subsec:infra_question_templates}
We compose user queries with two parts : (1) the \textbf{entity} of interest; and (2) the \textbf{task} to be executed. The base entities are \textit{Fund \textbf{X}}, \textit{Trust \textbf{X}}, and \textit{Custodian \textbf{X}}. We also combine the base entities, e.g. \textit{Fund \textbf{X}} and \textit{Custodian \textbf{Y}} to find the contract for this particular relationship. The possible tasks to apply on the entities include: \textbf{(1)} Explore all contracts; \textbf{(2)} Find \texttt{\{master agreements, master dates, termination dates, parties, clause X\}}; \textbf{(3)} \texttt{\{Summarize, Compare\}} clause X. These tasks are motivated by legal use cases.

\paragraph{Caching}
\label{subsec:infra_caching}
To reduce runtime latency, our system batch pre-processes data extraction.
This includes features related to the involved parties or dates. 
The extracted data is stored in a CSV file on the backend disk, acting as a cache. 
This cached data helps avoid latency especially for multi-step reasoning.
\paragraph{Section Search}
\label{subsec:infra_opensearch}
The large volume of legal text cannot be directly stored in our cache when retrieving contract sections.
Instead, our contracts are segmented and indexed in an OpenSearch distributed datastore provided by AWS. 
For each contract, we retrieve the top 20 most relevant sections using the BM25 ranking algorithm. The ranking algorithm looks at the presence of the target clause in both the section texts as well as its indexed title~(\S\ref{sec:section_title}).

\paragraph{Code Generation Agent}
\label{subsec:code_genagent}
Our system prompts the agent to generate Python code that can resolve the user's query 
(\S\ref{subsec:infra_question_templates}). 
The prompt includes tool names, descriptions, and examples similar to Chain-of-Thought~\cite{cot} and Self-Refine~\cite{selfrefine}. 
The prompt specifies instruction preferences, such as outputs to display for particular tools, and execution preferences such as not printing outputs or leaving incomplete \texttt{todo} tags.
\texttt{LAW} employs generated a three-tier system to generate and validate code: 
\begin{itemize}[noitemsep, leftmargin=*, topsep=0pt, partopsep=0pt, label={\tiny\raisebox{0.5ex}{$\blacktriangleright$}}]
    \item Syntax Validation: Performs pre-execution checks to verify code syntax, types, and security constraints.
    \item Hallucination Detection: Ensures generated code only calls tools that exist in \texttt{LAW}'s toolset with valid parameter signatures.
    \item Runtime Validation: Implements specialized error handling that captures and categorizes execution failures for targeted remediation.
\end{itemize} 
This verification framework enables \texttt{LAW}'s orchestrator to maintain a feedback loop, providing specific correction suggestions to the code generation agent when errors occur.

\begin{table}[!t]
\begin{tabular}{l|r|r}
\hline
\textbf{User Query}
& \multicolumn{1}{c|}{\textbf{\texttt{LAW}}}
& \multicolumn{1}{c}{\textbf{Baseline}}
\\
\hline
\multicolumn{1}{l}{\textit{Retrieval}} & \multicolumn{2}{c}{\textit{Hit Rate}} \\
\hline
Explore all contracts & 94.4 & 71.8\\
Find master agreements & 100.0 & 65.4\\
Find master dates & 93.3 & 36.2\\
Find termination dates & 95.4 & 2.5\\
Find parties & 100.0 & 16.3 \\
\hline
\multicolumn{1}{l}{\textit{Analytical}} &  \multicolumn{2}{c}{\textit{BERTScore F1}} \\
\hline
Summarize clause X & 89.5 & 68.1 \\
Compare clause X & 71.9 & - \\
\hline
\end{tabular}
\caption{A comparison of \texttt{LAW} with a simulated \texttt{gpt-3.5-turbo} baseline. For retrieval-type queries we measure the hit rate/recall calculating the percentage of correct retrievals compared to the ground truth. For analytical-type questions, we measure text similarity using BERTScore's~\citep{bert-score} F1 metric. The contextual embeddings for BERTScore are obtained using the \texttt{bert-large-uncased} model.}
\label{tab:results_1}
\end{table}

\section{Experiments}

\paragraph{Dataset Curation}
We labeled a dataset of 720 user queries as described in \S\ref{sec:eng_infra}. 
The tasks can be divided into two types: \textit{retrieval} and \textit{analytical}. \textit{Retrieval} queries correspond to retrieving information about entities of interest from contracts. Queries that involve the exploration of all contracts, the extraction of dates, and parties fall into this category. \textit{Analytical} queries require deeper insight, going beyond what can be extracted directly in the contracts. Queries that involve summarizing or comparing clauses across different contracts pertain to this category.
We generated 20 queries for each combination of task and entity for \textit{retrieval} queries and 10 for \textit{analytical} queries. We randomly populated the entities of interest from the universe of '40Act funds. 
The ground truth answers are generated using hand-coded scripts that leverage the same tools and text agents that the proposed system has access to. This procedure makes the evaluation agnostic to the implementation of the tools and focuses exclusively on \texttt{LAW}'s ability to generate code that correctly orchestrates the tools and the text agents. 

\paragraph{Baseline setup}
Our baseline seeks to understand if \texttt{gpt-3.5-turbo}, as is, can be prompted to answer queries on contracts. For \texttt{Explore all contracts} and \texttt{Find master agreement} queries, we simulate a noisy RAG framework by providing a set of four correct contracts and four distractor contracts. 
We reformulate user queries into a set of sequential True/False scenarios where the goal of the baseline is to determine if the candidate contract is associated with the entity, or is a master agreement, respectively. This choice was implemented as most contracts exceed the context limit of \texttt{gpt-3.5-turbo}.
For other queries, we choose four relevant contracts and prompt \texttt{gpt-3.5-turbo} to extract the desired pieces of information.
In essence, we narrowed the search space and provided relevant context for the baseline, where the provided context is sufficient for answering the queries. The context limit adds significant constraint on being able to provide in-context examples as demonstrations.

\paragraph{Results}
Table~\ref{tab:results_1} compares \texttt{LAW} against the baseline. \texttt{LAW} shows remarkable performance across \textit{retrieval} to \textit{analytical} queries.
Among \textit{retrieval} queries, for a true-false formulation of  \texttt{Explore all contracts}, the baseline performs reasonably at 71.8\% compared to 94.4\% achieved by \texttt{LAW}. This similar performance is seen for \texttt{Find master agreements}. The baseline starts performing poorly at 36.2\% when asked to lift the master date in contracts with a variety of dates. Moreover, the baseline quickly deteriorates for queries that require multi-hop reasoning, such as \texttt{Find termination dates}, where \texttt{LAW} surpasses the baseline by 92.9\%. We observe that the baseline tends to hallucinate immensely, showing a near-compulsion to conjure fictional dates. 
Specifically, this operation depends on the LLM 
finding the term for the duration of the contract and adding it to the master's effective date.
Finally, when asked to extract parties, the baseline often uses the question as an entity hint, but fails at lifting the complete list of entities from a dense agreement. For \textit{analytical} queries, the baseline underperforms on clause summarization because of an inability to understand which sections are relevant to a given clause. \texttt{Compare clause} requires understanding the trend of how a clause changed across multiple contracts.
This capability of comparing clauses cannot be performed using the baseline setup, as an agentic workflow with a larger context length is required.

\section{Conclusion}
We present \texttt{LAW}, a novel legal agentic workflow, achieving the successful completion of complex legal tasks. In contrast to fine-tuning an open-source LLM, our agentic invention is applicable for both closed and open-source models, leverages legal domain-specific tools and text agents that are modifiable and reusable, and orchestrates comprehensive plans. 
\texttt{LAW} has achieved remarkable performance, demonstrating robustness across \textit{retrieval} and \textit{analytical} queries and out-performing the baseline. Thus, our framework successfully enables automated workflows for varying contracts that govern the critical custody business. 
Future work can focus on applying \texttt{LAW} to non-English contracts, and explore additional agents grounded in other specific domains.  

\section*{Disclaimer}
This paper was prepared for informational purposes by the Artificial Intelligence Research group of JPMorgan Chase \& Co. and its affiliates ("JPMorgan'') and is not a product of the Research Department of JPMorgan. JPMorgan makes no representation and warranty whatsoever and disclaims all liability, for the completeness, accuracy or reliability of the information contained herein. This document is not intended as investment research or investment advice, or a recommendation, offer or solicitation for the purchase or sale of any security, financial instrument, financial product or service, or to be used in any way for evaluating the merits of participating in any transaction, and shall not constitute a solicitation under any jurisdiction or to any person, if such solicitation under such jurisdiction or to such person would be unlawful.

\bibliography{custom}
\clearpage
\appendix

\section{Domain Details}
\label{app:understanding_contracts}

\subsection{Form 485BPOS Regulatory Context}
Form 485BPOS is a 
document in the investment company industry, specifically used by mutual funds and other registered investment companies. It is filed with the U.S. Securities and Exchange Commission (SEC) under the Investment Company Act of 1940, often referred to as the '40 Act. It serves several key functions:
\begin{enumerate}[noitemsep, leftmargin=*, topsep=0pt, partopsep=0pt]
    \item \textbf{Registration}: It is used to register new mutual funds or update existing registrations.
    \item \textbf{Prospectus Updates}: '40Act Funds use Form 485BPOS to file updated prospectuses, which contain essential information for investors about the fund's investment objectives, risks, performance, and fees.
    \item \textbf{Regulatory Compliance}: It ensures that funds comply with SEC disclosure requirements and regulations subject under the Investment Company Act of 1940.
\end{enumerate}

Filing 485BPOS transcribes routine annual updates and other changes that become effective immediately upon filing. Currently, Form 485BPOS is 1.2\% of the total available EDGAR filings\footnote{\url{https://www.sec.gov/about/dera_edgarfilingcounts}}.

\subsection{Motivation and Impact}
The contracts analyzed by \texttt{LAW} govern critical relationships between custodian banks and '40Act funds, which are investment vehicles for trillions of retail investors, especially for retirement income. Proper governance of key clauses is paramount for the health of these funds, custodian banks, and the broader financial ecosystem. Traditionally, an immeasurable amount of time has been spent on finding, analyzing, and comparing key clauses in these contracts.

\texttt{LAW} shows potential for application to other contract types and legal documents. It also holds relevance for non-U.S. funds (e.g., USCITs) operating under similar regulations, demonstrating the broad applicability of our approach.

\subsection{Dataset Examples}
485BPOS Contracts in EDGAR are highly variable in length, format, content, and type. These include master agreements, amendments, separate appendixes, or the list of funds or entities that are captured by an agreement. See below for several example filings used in our dataset.

\begin{itemize}[noitemsep, leftmargin=*, topsep=0pt, partopsep=0pt, label={\tiny\raisebox{0.5ex}{$\blacktriangleright$}}]
    \item\textbf{Full Contract:} \url{https://www.sec.gov/Archives/edgar/data/1831313/000182912624004293/tcwetftrust_exg2.htm}
    \item\textbf{Single Amended:} \url{https://www.sec.gov/Archives/edgar/data/1879238/000182912623004720/bondbloxxetf_exg4.htm}
    \item\textbf{13th Amendment}: \url{https://www.sec.gov/Archives/edgar/data/1592900/000182912623003816/easeriestrust_ex99g1xiv.htm}
    \item\textbf{List of Funds:} \url{https://www.sec.gov/Archives/edgar/data/837274/000119312507144235/dex99gx.htm}
\end{itemize}

\subsection{Domain Complexity}
Analyzing lengthy legal contracts is difficult as there are no obvious headings, a plethora of legal concepts that are exceedingly difficult to digest, and no obvious categorization of paragraphs. These highly unstructured and dense legal documents encompass and describe in minute detail with different nuances in different formats the following contractual principles - such as \texttt{standard of care regimes}, \texttt{gross negligence}, \texttt{fiduciary responsibilities}, \texttt{breach of contract}, \texttt{liability for direct damages}, etc. These principles are presented in dense legal language, unstructured streams of text in different formats. Therefore, retrieving accurate numerical or textual values and analyzing/comparing them across tens of thousands of documents is a highly complex task. Within the legal domain, these retrieval and analytical tasks constitute as one of the most sophisticated and time-consuming tasks, especially in a highly unstructured database such as EDGAR, where no structured labels exist for the contracts.

\section{List of Key Clauses}
\label{app:clauses}

\begin{table}[!ht]
\begin{tabular}{l}
\hline
\textbf{Section Titles}                                                    \\ \hline
account transactions-                                                      \\ 
authorized persons                                                         \\ 
definitions                                                                \\ 
duties and responsibilities                                                \\ 
evidence of authority                                                      \\ 
fee schedule                                                               \\ 
fees and expenses                                                          \\ 
foreign custodian and subcustodian                                         \\ 
governing law                                                              \\ 
indemnification                                                            \\ 
instructions                                                               \\ 
limitations and scope of use or liability \\ 
miscellaneous                                                              \\ 
nominees                                                                   \\ 
proprietary information                                                    \\ 
recitals                                                                   \\ 
standard of care liabilities                                               \\ 
subcustodians and securities depositories \\ 
successor custodian                                                        \\ 
termination                                                                \\ \hline
\end{tabular}
\caption{List of key clauses found in fund custody contracts.}
\label{tab:key-clauses}
\end{table}

A full list of key clauses is found in Table~\ref{tab:key-clauses}.
The term \textit{Indemnification} refers to protective clauses that govern when losses occur with a third party. In the complex legal and financial landscape, recovery and punitive measures when accidents or losses occur is extremely important - for example, if a third party vendor's software breaks, is it the client or the provider's responsibility to recuperate those losses. Clauses governing \textit{Force majeure} events are often related to indemnification clauses, which refer to an event that is outside of a party's control and prevents them from fulfilling their obligations.
\textit{Termination} entails the date and conditions at which the contract/legal responsibility will end.

\section{System Design \& Implementation}
\paragraph{Framework} Our system is based on FlowMind's \citep{10.1145/3604237.3626908} framework and code recipe, extending it to a robust agentic legal framework with Fund Custody Services specific APIs. The code generation agent is responsible for mimicking planning by generating a series of function calls, akin to thinking steps, that breaks the question into steps semantically linked to our tool calls.

\paragraph{Tools and Agents} \texttt{LAW} incorporates tools for direct extraction, multi-hop reasoning, and text analysis. By allowing \texttt{LAW} the flexibility to reuse statements, pass previous information from a function call directly into another, and iterate over retrieved items, it can go beyond single-step reasoning. Text-based agents employ zero-shot prompting. The summarize and compare tools utilize specialized text agents to yield useful analytics, enhancing the system's capability to handle complex queries. Specifically, the summary agent's task is to succinctly summarize a particular clause, restricting output to facts present in the text, such as preserving key terms, entities, dates, etc.

\paragraph{Long-Context Contracts} To handle long clauses and contracts, our system breaks the text into smaller chunks. These are then processed by separate "sub-agent" spawns to summarize, after which the results are concatenated into a complete summary. This approach ensures comprehensive analysis of extensive legal texts that would not fit into a standard context window.

\paragraph{Framework Extensibility}
While \texttt{LAW} currently demonstrates strong performance with its existing toolset, extending the framework to new tasks requires careful consideration. Adding new tools involves:
\begin{itemize}[noitemsep, leftmargin=*, topsep=0pt, partopsep=0pt, label={\tiny\raisebox{0.5ex}{$\blacktriangleright$}}]
    \item Task Analysis: Identifying atomic operations that can be modularized into reusable components.
    \item Tool Development: Creating focused tools with clear inputs/outputs and comprehensive unit tests.
    \item Integration Testing: Verifying the tool's interaction with the code generation agent and other components.
\end{itemize}
The modular nature of \texttt{LAW} allows new tools to be added without modifying existing components. However, several challenges we faced included:
\begin{itemize}[noitemsep, leftmargin=*, topsep=0pt, partopsep=0pt, label={\tiny\raisebox{0.5ex}{$\blacktriangleright$}}]
    \item Ensuring tool reliability across diverse contract formats.
    \item Maintaining clear boundaries between tool responsibilities.
    \item Balancing tool specificity with reusability.
    \item Managing increased complexity in the orchestration logic.
\end{itemize}

\section{Experimental Details}
\subsection{Dataset Creation and Validation}
We collaborated with business end-users to identify useful pieces of information to extract and sections/titles that they often examine. For validation, we conducted a small pilot study with human users on shorter contracts, obtaining performance comparable to that reported in the paper.

\subsection{Evaluation}
Our evaluation methodology included comparing outputs against hard-coded scripts as a ground truth. We simulated a noisy RAG framework by obtaining a small subset of relevant and irrelevant documents, mimicking a scaled version of the 485BPOS dataset. In this framework, we evaluate the system's ability to ignore irrelevant contracts while effectively extracting information from the relevant ones. The experiments employed a few-shot prompting strategy.

\subsection{Error Analysis}
The main source of error in obtaining termination dates was the hallucination of non-existent dates, which particularly affected baseline performance. We noted that generating baseline output for clause changes across contracts was not feasible due to context limitations when communicating pairs of clauses.

\section{Potential \texttt{LAW} Queries}
\begin{itemize}[noitemsep, leftmargin=*, topsep=0pt, partopsep=0pt, label={\tiny\raisebox{0.5ex}{$\blacktriangleright$}}]
    \item Find the termination dates of all contracts from custodian Goldman Sachs.
    \item Find the master dates of contracts between Trust Investor Counselor Series Fund Inc and Custodian State Street Bank and Trust.
    \item Compare the fees and expenses clause of the previous contracts.
\end{itemize}

\section{Title and Section Retrieval Metrics}\label{app:section_accuracy}

\begin{table}[h!]
    \centering
    \small
    \begin{tabular}{l|c|c}
         \hline
\text{Attributes}            & \texttt{T5-Base}    & \texttt{T5-Large} \\ \hline
\text{Method}                & \text{Generation} & \text{Classification} \\
\text{Parameters}            & 222,903,552       &  737,668,096 \\
\text{Samples / Title} & 400               & 1500 \\
\text{Accuracy}              & 42\%              & 46\% \\
\text{GPU}                   & \text{NVIDIA T4}  & \text{NVIDIA A10G} \\
\text{\# of GPUs}        & 4        & 4 \\
\text{GPU Memory}            & \text{64 GiB}        & \text{96 GiB} \\ \hline
    \end{tabular}
    \caption{Experiment and environment details for section title generation and classification. }
    \label{tab:sectionenv}
\end{table}

\noindent
The following are the training parameters for \texttt{T5-Base} and \texttt{T5-Large} models.

\begin{itemize}[noitemsep, leftmargin=*, topsep=0pt, partopsep=0pt, label={\tiny\raisebox{0.5ex}{$\blacktriangleright$}}]
    \item Training epochs: 3
    \item Per device batch size for train and eval: 2
    \item Learning rate: 0.01
\end{itemize}

\section{Sample Generated Code}
\label{subsec:generated_code}

\textbf{Query:} Compare the \textit{authorized persons} clauses for Fund \textit{BNY Mellon International Equity Income Fund}.
\newline
\textbf{Customized suffix:} Only compare subsequent clauses of five sampled non-empty contract sections. Ensure that there are also contracts for this entity choice.
\newline
\textbf{Code generated:}
\begin{lstlisting}[language=python]
# Step 1: Retrieve all agreements for the specified fund.
agreements = get_agreements_for(funds='BNY Mellon International Equity Income Fund')

# Step 2: Check if any agreements were found.
if agreements:
    # Step 3: Extract the clauses for the specified section from each agreement.
    clauses, urls, _ = get_section_v2(agg_list=[agreement[1] for agreement in agreements], section_name='authorized persons')

    # Step 4: Check if any clauses were found.
    if clauses:
        non_empty_clauses = [clause for clause in clauses if clause != '']
        num_clauses = len(non_empty_clauses)
        if num_clauses > 5:
            interval = num_clauses // 5
            selected_clauses = non_empty_clauses[::interval]
        else:
            selected_clauses = non_empty_clauses

        # Step 5: Compare the selected clauses.
        output = get_comparison_v1(
        list_agreement_tuples=[(agreement[0], agreement[1]) for agreement in agreements], text_list=selected_clauses)
    else:
        output = "No 'authorized persons' clauses found for Fund 'BNY Mellon International Equity Income Fund'"
else:
    output = "No agreements found for Fund 'BNY Mellon International Equity Income Fund'"

# Step 6: Return the output.
output
\end{lstlisting}

\end{document}